\documentclass[letterpaper, 10 pt, conference]{ieeeconf}  
\newcommand{\myheader}{Preprint. Accepted as a conference paper at the \textit{International Conference on Intelligent Robots and Systems} (IROS) 2018, \textcopyright IEEE 2018}

\usepackage{atbegshi,picture}

\usepackage{datetime}            

\usepackage{lastpage}            

\newcommand{\myleftstd}{1.0in}

\ifthenelse{\isodd{\thepage}}
 {\newcommand{\myleftmargin}{\oddsidemargin+\myleftstd}}
 {\newcommand{\myleftmargin}{\evensidemargin+\myleftstd}}

\AtBeginShipoutNext { 
 \AtBeginShipoutUpperLeft{%
  \put(\dimexpr 0.2\myleftmargin\relax,-1.2cm){\parbox{\textwidth}{\footnotesize\sffamily\noindent{\centering \myheader \hfill}}}%
}}

\IEEEoverridecommandlockouts                              
\usepackage{graphicx}
\usepackage{xcolor}
\colorlet{darkblue}{blue!50!black}
\colorlet{hlinkcolor}{darkblue}
\makeatletter
\let\NAT@parse\undefined
\makeatother
\usepackage[numbers]{natbib}

\newcommand{\mb}[1]{\ensuremath{\mathbf{#1}}}

\title{\LARGE \bf
Deep Neural Object Analysis by Interactive Auditory Exploration with a Humanoid Robot
}

\author{Manfred Eppe$^{1}$, Matthias Kerzel$^{1}$, Erik Strahl$^{1}$ and Stefan Wermter$^{1}$
\thanks{*The authors gratefully acknowledge partial support from the German Research Foundation DFG under project CML (TRR 169) and the Hamburg Landesforschungsf\"orderungsprojekt CROSS.}
\thanks{$^{1}$Knowledge Technology, Department of Informatics, University of Hamburg, Germany.
        {\tt\small \{eppe, kerzel, strahl, wermter\} @informatik.uni-hamburg.de}}%
}

\begin{document}
\maketitle
\thispagestyle{empty}
\pagestyle{empty}

\begin{abstract}
We present a novel approach for interactive auditory object analysis with a humanoid robot. The robot elicits sensory information by physically shaking visually indistinguishable plastic capsules. It gathers the resulting audio signals from microphones that are embedded into the robotic ears. A neural network architecture learns from these signals to analyze properties of the contents of the containers. Specifically, we evaluate the material classification and weight prediction accuracy and demonstrate that the framework is fairly robust to acoustic real-world noise. 
\end{abstract}

\section{INTRODUCTION}
\label{sec:introduction}

Currently, several humanoid companion robots are developed to act alongside humans in complex multimodal domestic environments that are designed to cater to human sensory and motor abilities (e.g.~\cite{kerzel2017nico}, \cite{Oliveira2012}, \cite{Eppe2016}, \cite{Eppe2018CNN}). Therefore, companion robots should be able to utilize human strategies for perception tasks to cope with perceptual ambiguities.
Specifically, robots should be capable of performing \emph{interactive perception}, i.e. they should be capable of executing actions for the purpose of creating perceivable signals that they can process to gather useful information about their environment. 

Interactive perception is a common human exploratory strategy that is useful if passive perception does not yield a specific desired information. For example, when humans are unable to distinguish objects visually, they interact with the objects and focus on signals from other sensory channels that result from the interaction. For instance, shaking a hollow object elicits haptic and auditory sensations that can be used to determine if the object is empty or filled, what material it is filled with, and how much of the material it contains.

%
%
In this paper, we focus on the auditory sense and extend the state of the art in interactive audio perception by presenting an embodied neural architecture that performs interactive exploration with a set of visually indistinguishable objects and learns to identify individual object materials and weights. Specifically, we address the following research questions:

\begin{enumerate}
\item Can interactive auditory exploration be used for the analysis of visually indistinguishable objects, specifically for  classifying and weighting materials?
\item How robust is auditory object analysis with respect to external sound sources and noise?
\end{enumerate}

\begin{figure}
\centering
\includegraphics[width=0.37\textwidth]{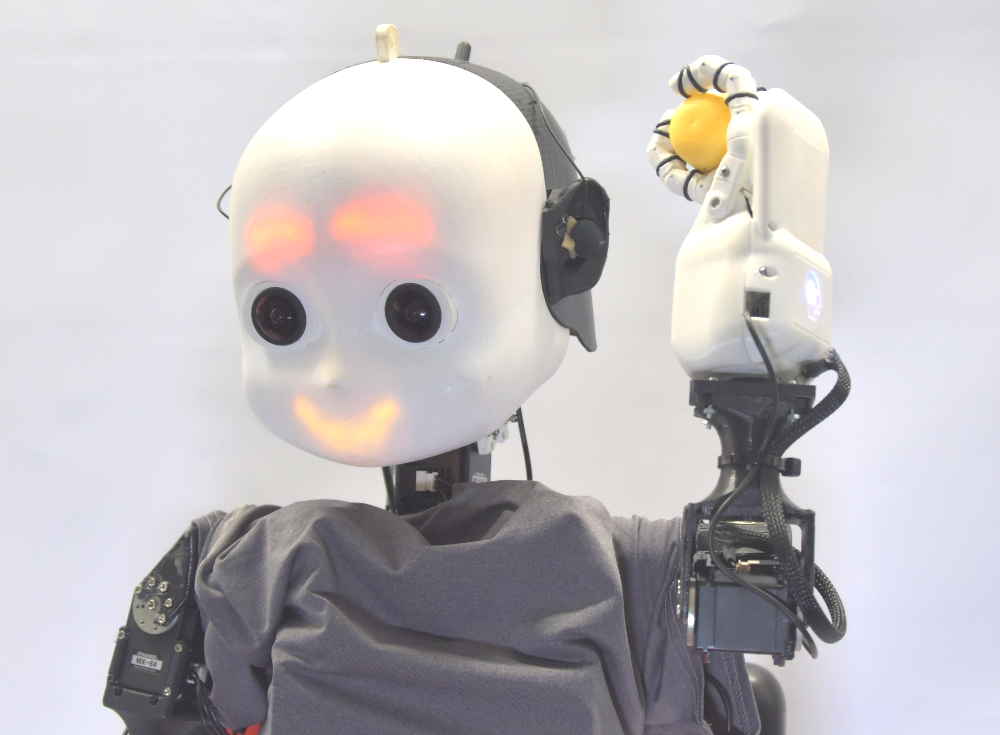}
\caption{The NICO robot is shaking a plastic capsule to learn about its content via auditory signals obtained from the microphones embedded in NICO's realistic pinnae.}
\vspace{-10pt}
\label{fig:nico_shaking}
\end{figure}

\begin{figure*}[ht]
\centering
\includegraphics[width=.8\textwidth]{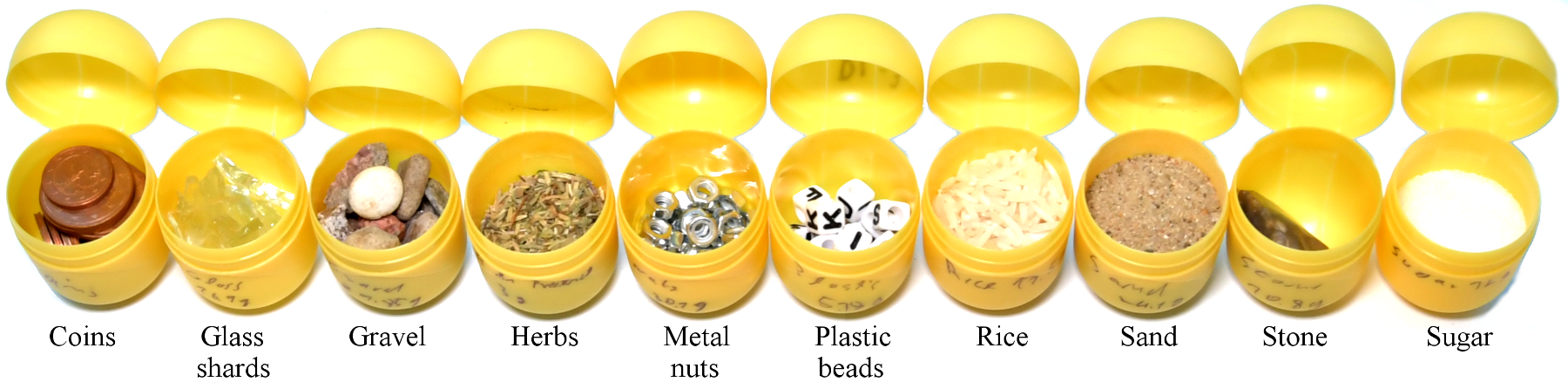}
\vspace{-5pt}
\caption{The sample capsules with 10 different materials.}
\vspace{-10pt}
\label{fig:capsules}
\end{figure*}

To address the active nature of explorative perception, the neural model is embodied in NICO, the \underline{N}euro-\underline{I}nspired \underline{CO}mpanion robot \cite{kerzel2017nico}. 
NICO is an adaptable humanoid research platform,
equipped with the motor skills required to interact with, grasp and manipulate objects \cite{Eppe2017}. Its auditory perception is realized by embedding stereo microphones into realistic pinnae (see Fig.~\ref{fig:nico_shaking}). To answer the research questions of this work, we conduct experiments where NICO uses its hand and arm to actively shake capsules filled with different amounts of different bulk materials, such as sand, rice, glass shards, etc. (see Fig.~\ref{fig:capsules}).%
\footnote{An explanatory video has been submitted as an attachment to this paper, and we also present an edutainment version that involves robotic interactive auditory perception \cite{Strahl2018}.}
To process the audio signals that emanate from the shaking, we implement a recurrent neural network (RNN) architecture that performs classification and regression for interactive material analysis. 

\section{BACKGROUND AND RELATED WORK}
\label{sec:related_work}



\subsection{Interactive Auditory Perception in Robotics}
\label{sec:robot_audio_sensing}
Interactive perception is commonly described as a search over a structured space $\mb{S} \times \mb{A} \times \mb{T}$ (sensing, action, time) (e.g. \cite{Martin-Martin2016}). Perception is passive if $\mb{A}=\emptyset$, and  active or interactive if $\mb{A} \neq \emptyset$ \cite{Bohg2017}. In signal processing and Artificial Intelligence research, there exists a large body of studies in passive auditory perception, e.g., in speech recognition, or human emotion recognition \cite{Barros2016}. 

For active perception, $\mb{A}$ consists of non-forceful actions that affect only the robot's sensor properties. For example, active vision usually refers to the active moving of a camera with the purpose to observe a scene from a different angle, so that, e.g., more salient parts of a scene become visible \cite{Bellotto2012}.
Active perception can also be executed in a cross-modal manner, e.g., in cases where data from one sensor provides clues about how to use another sensor \cite{Martin-Martin2017}.

Robots, however, are embodied agents and can go beyond passive and active perception. Like humans, they can manipulate objects in the environment to perform \emph{interactive perception}, i.e., behavior where $\mb{A}$ consists of forceful actions that manipulate objects. 
Abstractly speaking, interactive perception allows an embodied agent to deliberately perform an action, perceive the effect of the action and infer information about the conditions under which the action causes a particular sensation. In classical AI and action theory, this inference is referred to as \emph{postdiction} \cite{Eppe2015e,Eppe2013c}. 
The interactive perception and postdiction capability allows agents to solve various problems including manipulation skill learning, object recognition, object dynamics learning and state representation learning \cite{Bohg2017}. 

Even though humans frequently use interactive auditory perception (e.g. blind people performing echolocation, craftsmen knocking on material to determine its quality, or children learning to play a musical instrument), there is only little work on robots performing interactive auditory perception. 
An early approach that involves interactive sound perception has been realized by \citet{Torres-Jara2005}, who have a robot tapping on four different objects of different materials and identifying them using auditory and tactile perception by manual analysis of the resulting data.
Another example where active auditory perception is performed is the work by \citet{Natale2007}, who realize a baby-like robot that learns sensorimotor coordination by interaction with the world, using multimodal sensory feedback including acoustic signals. 
The work that is probably most related to ours has been conducted by \citet{Sinapov2014,Sinapov2014a}. The authors use a multimodal approach that includes interactive auditory perception to distinguish 100 different objects \cite{Sinapov2013,Sinapov2014a}, and they also perform classification of bulk materials inside hollow capsules \cite{Sinapov2014}. They do not only classify unary object properties, such as material, weight class and color, but also binary relations between objects. Their robot performs 10 different interaction behaviors, including shaking, lifting and poking of the capsules, and employs a C4.5 algorithm to perform classification. This way, the authors achieve accuracy rates above 95\% for material, color and weight classification. However, \citet{Sinapov2014} use only four material types, specifically glass, screws, beans and rice, which have more distinctive sound characteristics than our ten different materials, which include, e.g., sand and sugar (see Fig.~\ref{fig:capsules}). Furthermore, the authors classify weight to distinguish between light, medium and heavy objects, whereas in our work we predict the actual weight in grams using a linear regression model.


\subsection{The humanoid robot NICO}
To realize the interactive perception, we employ NICO, the Neuro-Inspired COmpanion \cite{kerzel2017nico} robot. NICO is a humanoid flexible and modular research platform for crossmodal learning, neuro-cognitive modeling and multimodal  human-robot  interaction. Its open and freely available design\footnote{Visit \url{http://nico.knowledge-technology.info} for more info and video material.} allows one to modify it to meet the specific requirements of different experimental setups. With a height of 101 cm, NICO has a child-like form. It has two arms with 6 degrees of freedom that have a human-like range of motion which enables it to carry out the auditory exploration procedures.

To this end, NICO grasps the sample capsules with a three-fingered SR-DH4D hand\footnote{\url{http://www.seedrobotics.com/}, accessed 25th Feb.~2018}. The three-jointed fingers of the hand are tendon-operated and curl around smaller objects to ensure a stable grasp without sophisticated grasp planning or sensory feedback. 

NICO perceives auditory information via two Soundman OKM II binaural microphones that are embedded in human-shaped and 3D-printed pinnae. Though primarily designed to study human-like binaural hearing for sound source localization and speech recognition, this distribution and embedding of sensors allow for a good adaptation of human-like audio-tactile exploratory procedures. NICO's head was designed not to contain any sound source (e.g., cooling device) whose ego noise would be detrimental to the experimental setup. However, since the auditory exploratory procedures are active motor actions, is unavoidable that ego noise from the gears of the servomotors of the robotic arm is produced during data recording.










\section{NEURAL SIGNAL PROCESSING ARCHITECTURE}
\label{sec:signalproc}
\begin{figure}
\centering
\includegraphics[width=0.4\textwidth]{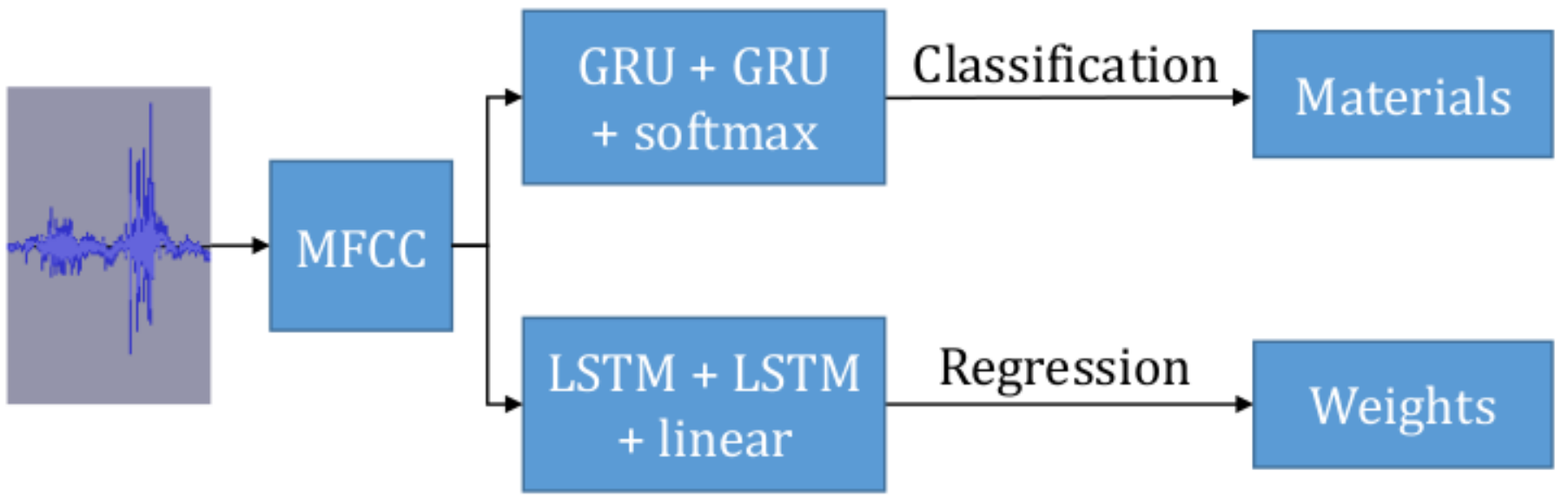}
\vspace{-8pt}
\caption{Neural architecture for classification and regression using audio information.}
\vspace{-12pt}
\label{fig:architecture}
\end{figure}

For analyzing the audio signals, we first employ Mel Frequency Cepstral Coefficients (MFCC) preprocessing to generate a frequency spectrum in a neurocognitively plausible way (see Sec.~\ref{sec:signalproc:preproc}). The resulting spectral signals are fed into two separate neural networks to perform material classification and weight prediction (see Fig.~\ref{fig:architecture}, Sec.~\ref{sec:signalproc:nn}). We obtained hyperparameters for the preprocessing and network architecture by performing hyperparameter optimization using a tree-structured Parzen estimator (TPE) \cite{Bergstra2013}. 

\subsection{Auditory Preprocessing}
\label{sec:signalproc:preproc}
Human auditory perception is based on hair cells within the cochlea that respond to certain frequencies. Higher frequencies are transmitted by hair cells at the beginning of the cochlea, while low frequencies are transmitted by hair cells near the end of the cochlea. This biological realization resembles a spectral frequency analysis, where different hair cells correspond to different frequency bands. A mathematical and neurophysiologically plausible approach to realize this behavior computationally is to use the Mel Frequency Cepstrum (MFC). MFC analysis first performs a Fourier transform of a sliding signal window, mapping powers of the spectrum onto the Mel scale. It then performs a discrete cosine transform of the logs of the powers and returns the Mel Frequency Cepstral Coefficients (MFCC), i.e., the amplitudes of the resulting spectrum. 

We tested different signal window sizes and step sizes and found that these parameters did not significantly affect the classification and regression performance as long as the window size is between 15ms and 50ms, and the step size is below the window size. For our setup, we chose 30ms as window size and 15ms as window step size. 
The frequency spectrum was split into 21 Mel coefficients for material classification and 27 Mel coefficients for weighting, as these valued yielded the best results during hyperparameter optimization. We also normalized the sound input to 0db.

\subsection{Neural Architecture}
\label{sec:signalproc:nn}
Recurrent Neural Networks (RNN) are well-known for their capability to perform classification and regression tasks with a continuous variable-length input, e.g., language or audio processing \cite{Eppe2018ICANN,Barros2016,Alpay2016,Hinaut2015a}. 
The ability to perform continuous audio processing of variable-length samples is important for active exploration because it provides the system with a high exploration flexibility. For example, for some samples it may be sufficient to shake them only once until a classification with high confidence is achieved, while for others the robot may need to shake the samples more often and at different angles, which would result in longer audio sequences. We have not yet implemented such dynamic exploration behavior, but we used Recurrent Networks already at this stage of our work as a preparation to realize respective experiments in the future. 

Specifically, we experimented with Simple Recurrent Networks (SRN), Long Short-Term Memory (LSTM) \cite{Hochreiter1997}, and Gated Recurrent Units (GRU) \cite{Chung2014}. LSTM and GRU have the advantage of being gated architectures, which are significantly less prone to the vanishing or exploding gradient problem \cite{Hochreiter1997}, and therefore more suitable for backpropagation. This advantage has been reflected during our hyperparameterization experiments, which showed the LSTM and GRU perform significantly better than SRN. 
%
%
We found that two recurrent LSTM or GRU layers, where the first layer consists of around 300 to 700 units and the second of around 50 to 100 units, worked best for both classification and regression. For both tasks, we used the best layout as obtained from the hyperparameter optimization. Specifically, for the classification experiments, we used a 491 unit GRU followed by a 99 unit GRU, followed by a single dense softmax layer. For the weight regression, we used a 376 unit LSTM followed by a 69 unit LSTM followed by a single dense linear layer. 
We also experimented with stacking more dense and recurrent layers, but this resulted in a decreased classification and regression accuracy. Adding a dropout layer did not have a measurable effect on the accuracy.

\section{EXPERIMENTAL SETUP}
\label{sec:experimental_setup}
In the presented experimental setup, the humanoid NICO explores a set of visually non-distinct plastic capsules (see Fig.~\ref{fig:capsules}). It grasps the capsules with its robotic hand and repeatedly performs an audio-tactile exploration procedure by shaking the objects near its ear, as depicted in Fig.~\ref{fig:nico_shaking}. 


\subsection{Auditory Exploration Procedure}
The capsules that NICO explores have a cylindrical shape with rounded top and bottom ends. They have a diameter of three centimeters and a length of five centimeters. 
The containers are filled with ten different materials in three different quantities, resulting in a test set of 30 different containers, as illustrated in Fig.~\ref{fig:capsules}. 
To elicit auditory information from the objects, NICO performs an exploration procedure by raising the grasped capsule to its ear and shaking it in a whipping motion. 
The whipping is generated by up-down shaking motions of NICO's elbow joint, at a frequency of approximately 1 Hz. 
This causes the content of the capsule to decelerate sharply, resulting in auditory rattling sensations. 
We implement two different exploration movements, in which the robot shakes the capsules in different directions by tilting the wrist motor by 90 degrees. This changes the direction of deceleration of the materials inside the containers and results in a more diverse dataset.

\begin{table*}[ht]
\centering
\vspace{5pt}
\caption{Auditory classification results as confusion matrix}
\label{tab:classification}
\begin{tabular}{r|llllllllll}
-- &	Coins &	Glass &	Gravel &	Herbs &	Nuts &	Plastic  &	Rice &	Sand &	Stone &	Sugar \\\hline 
Coins &	\textbf{0.94} &	0.02 &	0.01 &	0.00 &	0.00 &	0.00 &	0.00 &	0.00 &	0.03 &	0.00\\
Glass &	0.01 &	\textbf{0.91} &	0.02 &	0.00 &	0.05 &	0.00 &	0.00 &	0.00 &	0.01 &	0.00\\
Gravel &	0.03 &	0.13 &	\textbf{0.83} &	0.00 &	0.00 &	0.00 &	0.00 &	0.00 &	0.01 &	0.00\\
Herbs &	0.00 &	0.00 &	0.00 &	\textbf{0.88} &	0.00 &	0.00 &	0.00 &	0.04 &	0.00 &	0.08\\
Nuts &	0.00 &	0.01 &	0.00 &	0.00 &	\textbf{0.98} &	0.00 &	0.00 &	0.01 &	0.00 &	0.00\\
Plastic  &	0.00 &	0.01 &	0.00 &	0.00 &	0.00 &	\textbf{0.99} &	0.00 &	0.00 &	0.00 &	0.00\\
Rice &	0.00 &	0.00 &	0.00 &	0.00 &	0.00 &	0.00 &	\textbf{1.00} &	0.00 &	0.00 &	0.00\\
Sand &	0.00 &	0.00 &	0.00 &	0.01 &	0.00 &	0.00 &	0.01 &	\textbf{0.90} &	0.00 &	0.08\\
Stone &	0.00 &	0.00 &	0.02 &	0.00 &	0.00 &	0.00 &	0.00 &	0.00 &	\textbf{0.98} &	0.00\\
Sugar &	0.00 &	0.00 &	0.00 &	0.11 &	0.00 &	0.00 &	0.00 &	0.21 &	0.00 &	\textbf{0.69}\\
\end{tabular}
\end{table*}

\begin{table*}[ht]
\centering
\caption{Weight regression results for all materials}
\label{tab:weights}
\begin{tabular}{r|llllllllll|l}
Material  									&	Coins &	Glass &	Gravel &	Herbs &	Nuts &	Plastic &	Rice &	Sand &	Stone &	Sugar & avg.\\\hline 
Sample 1 weight  							&	20.7 &	6.3 &	10.4 &	1.0 &	9.9 	&	1.7 &	4.5 	&	8.0 &	4.5 &	4.0 & 7.10\\
Sample 2 weight  							&	39.1 &	12.6 &	20.4 &	2.0 &	20.1 	&	3.4 &	9.0 	&	16.0 &	7.3 &	8.0 & 13.79 \\
Sample 3 weight  							&	61.8 &	18.9 &	29.9 &	3.0 &	30.1 	&	5.2 &	13.5 	&	24.1 &	10.8 &	12.0 & 20.93 \\\hline
Mean sample weight							&	40.6 &	12.6 &	20.2 &	2.0 &	20.0 	&	3.4 &	9.0 	&	16.0 &	7.5 &	8.0 & 13.13 \\\hline
Mean prediction error [g]:  				&	11.01 &	3.16 &	4.45 &	2.09 &	3.93 &	1.19 &	2.13 &	4.31 &	3.12 &	2.81 & \textbf{3.51}\\
Mean prediction error [\% of mean weight]: 	& 27.12	& 25.08	& 22.02	& 104.5	& 19.51	& 34.73 & 23.67	& 26.88	& 41.42	& 35.12 & \textbf{36.01}\\
\end{tabular}
\vspace{-10pt}
\end{table*}

\subsection{Dataset Recording}
The datasets are recorded in an office environment, with realistic background noise. Two persons are working on computers in the same room, and additional noise is created from people walking and talking in the outside corridor. Two full sets of 10 trials per object are performed for both exploration procedures. We randomize the order of recording to ensure that no auditory clues from the background noise could be used to identify samples. Each sample is handed over by an experimenter to the robot. During the grasping, the experimenters ensure that the plastic capsule containing the sample is tightly grasped by the hand (see Fig.~\ref{fig:nico_shaking}). The audio is recorded at 48kHz stereo using both the left and right ear's microphone. Each of the 30 capsules is given to the robot two times, and for each recording, the robot shook the same capsule 18 times. This provides us with 36 auditory signals per capsule and 1080 signals in total. Each signal is trimmed to a length of 625ms. 


\subsection{Neural Network Training}

\begin{figure*}[ht]
\centering
\vspace{5pt}
\includegraphics[width=0.80\textwidth]{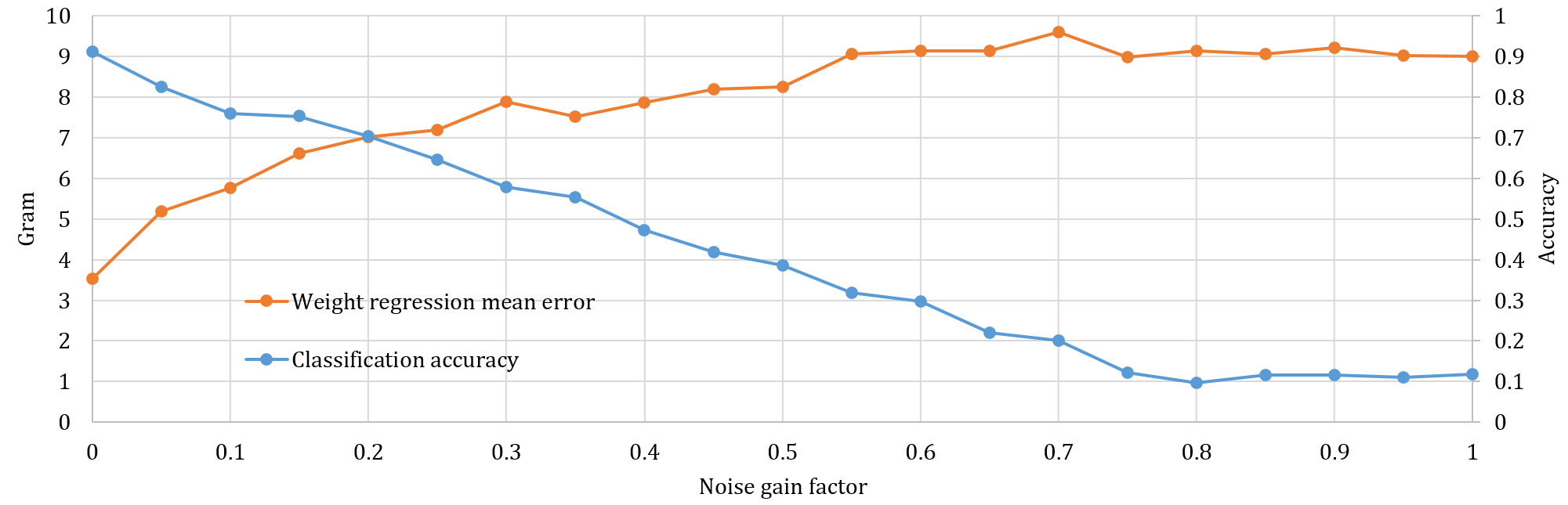}
\vspace{-5pt}
\caption{Auditory classification and weighting in the presence of background noise.}
\vspace{-5pt}
\label{fig:bg_noise}
\end{figure*}

For training the neural networks, we split the recorded data into 80 test and validation samples and 1000 training samples. We randomly generate the split 15 times, performing training and evaluation individually for each of the 15 splits. The results presented in Table~\ref{tag:classification}, Table~\ref{tag:weights} and Fig.~\ref{fig:bg_noise} represent the average over all 15 runs. We use a batch size of 16, and an early stopping mechanism stops the training process when the loss does not improve after two consecutive training and evaluation runs. 
We do not focus on training and prediction performance in terms of computation time but find that the system is well capable of real-time prediction. Training the network on 4 cores of an Intel Xeon CPU takes around 10 minutes for classification and 20 minutes regression. Prediction takes in the order of magnitude of 25ms per sample for both classification and regression.

\section{RESULTS}
We report the results for classifying the material and predicting the weight of a sample in Table~\ref{tag:classification} and Table~\ref{tag:weights} respectively. 
We demonstrate that our approach can perform both tasks reasonably well, and Fig.~\ref{fig:bg_noise} shows that it is also relatively robust to indoor noise. 

\subsection{Material classification}
\label{sec:res:mat_class}
To obtain the classification results, we compute a confusion matrix that results from averaging over all 15 train/predict sessions on randomly split datasets (see Table~\ref{tag:classification}). 
We find that rice was the most distinctive material, as it has been correctly classified in all test cases. 
Sugar, sand, and herbs are hardest to classify because they provide a very low acoustic signal and are often confused with each other. The classification performance of gravel is also below average because it is often confused with glass. There is relatively little confusion for all other cases, and the \emph{overall average classification performance is 91\%}. 
We did not perform rigorous baseline experiments with human subjects, but preliminary tests showed that especially the difficult cases, such as differentiating between sand and sugar, are very hard, if not impossible, for humans to perform reliably.

\subsection{Weight regression}
\label{sec:res:weight_reg}
The mean absolute error (MAE) when performing weight regression is 3.51g. As a baseline, consider that the mean weight of all capsules is 13.13g, and guessing this average in all cases, results in a mean error of 9.4g. 
In our experiments, we observe that the MAE is higher for heavier capsules, i.e, those that contain coins, gravel and sand. This relation between material weight and prediction error is, however, not strictly linear. Weight regression worked best for the capsules containing plastic beads, where the MAE is only 1.19g. 

Expressed in percentage of the mean capsule weight of the same material, the overall average prediction error is around 36\% in average over all classes. These results are not suitable for precise measurements, but allow a robot to perform fair estimates about weight, and, in combination with material classification, also about the volume of a material inside a container. 
This capability may be important for interaction scenarios where objects need to be grasped and handed over to other individuals, or where quantities of materials need to be estimated.

\subsection{Robustness to external noise}

The sound samples used in all experiments are already recorded under real-world conditions in an office environment. We took care that people in the office are not talking, but background noise like typing on a keyboard and people walking around are clearly identifiable on the sample data. 
In addition, there is a significant amount of ego noise coming from the robot's servos. Hence, the results depicted in the previous Sections \ref{sec:res:mat_class} and \ref{sec:res:weight_reg} already involve a realistic amount of noise. 
However, in order to make more precise statements about robustness to noise, we also perform experiments where we simulate an environment with other external sound sources at varying levels. Therefore, we use six randomly selected samples from different background noises including traffic, people speaking, airport, etc.\footnote{Noise samples obtained from \url{http://www.orangefreesounds.com/}, 3rd Jan. 2018.}, and overlay the noise signals with the robot's auditory signal before repeating training and prediction experiments. The gain factor of the normalized noise is increased by steps of 0.05, while the gain factor of the normalized audio signal from the shaking is decreased by steps of 0.05. The mixed signal field strength is kept constant at a normalized gain factor of 1. 
Our results are illustrated in Fig.~\ref{fig:bg_noise}. 
We observe that a classification accuracy above 75\% is still achieved for cases where the noise gain factor does not exceed 0.15. The mean absolute error (MAE) for weight regression rises from around 3.5g to around 6g if the noise gain factor rises from 0.0 to 0.1. 
Hence, we conclude that classification is more robust to noise than weight regression. However, assuming that indoor environments have a realistic noise gain level of approximately 0.1, and that the sound samples already involve an implicit basic amount of background office noise and robotic ego noise, our results show that both tasks perform sufficiently well under real-world indoor conditions.



\section{CONCLUSION}
We present a humanoid robot that performs auditory exploration procedures on a set of visually indistinguishable plastic containers filled with different amounts of various materials. We show that a deep recurrent neural architecture can learn to distinguish individual materials and also estimate their weights. Our work implies that robots that are equipped with proper audio recording devices are very capable of analyzing visually indistinguishable material samples, which answers our first research question. Our results apply to a real-world environment with the typical background noise of an office environment and the ego noise of the robot's servo motors. We show that both classification and regression work reasonably well under such realistic conditions. Our second research question is, therefore, also answered. 

To the best of our knowledge, there exists no other recent approach, besides the work by \citet{Sinapov2014}, where material is classified and analyzed by interactive robotic auditory perception. Our results are not directly comparable to those of \citeauthor{Sinapov2014} due to different experimental setups, and because the focus of \citet{Sinapov2014} is more on classifying binary object relations and on multimodal perception, but we have achieved a comparable fair classification and weight prediction accuracy while using more different materials that are harder to distinguish than those used by \citet{Sinapov2014}. 

With this research, we have contributed towards humanoid robots that can experience and learn in a complex multimodal environment by using interactive exploratory procedures that help to distinguish objects in the absence of visual cues. 
In our setup, we have performed independent learning experiments for weight regression and classification because the hyperparameter optimization results suggested using slightly different network architectures (GRU for classification and LSTM for regression). However, the results indicate that one could also use a combined recurrent architecture where only the final softmax (for classification) and linear (for regression) layers are trained independently. 

For future work, we will extend our approach to more complex scenarios where NICO has to explore a variety of everyday objects using also other senses, such as tactile information. We argue that a fusion of tactile and auditory signal processing will further improve the results, in particular under noisy conditions. We will also implement a more dynamic physical exploration procedure, where the result of the classification has an effect on the exploration. For example, a certain material may have salient acoustic characteristics when shaking at a certain angle or at a certain frequency. The interactive sensing mechanism that results from this feedback loop allows the robot to adapt its exploration dynamically to realize such behavior.

\bibliographystyle{IEEEtranN}
\bibliography{myRefs}{}



\end{document}